\documentclass[9pt]{elife}

\usepackage{lipsum} 
\usepackage[version=4]{mhchem}
\usepackage{siunitx}
\DeclareSIUnit\Molar{M}
\usepackage{pdflscape}
\usepackage{multicol}
\usepackage{graphicx}
\usepackage{adjustbox}
\usepackage{geometry}
\usepackage{makecell}
\usepackage{longtable}
\usepackage{rotating}
\usepackage{todonotes}

\usepackage{tikz}
\usetikzlibrary{fit,positioning}
\usetikzlibrary{mindmap}

\title{Moo-ving Beyond Tradition: Revolutionizing Cattle Behavioural Phenotyping with Pose Estimation Techniques}

\author[1,2]{Navid Ghassemi}
\author[3]{Ali Goldani}
\author[4]{Ian Q. Whishaw}
\author[2,4*]{Majid H. Mohajerani}
\affil[1]{McGill University Integrated Program in Neuroscience, Montréal, Québec, Canada}
\affil[2]{Department of Psychiatry, Douglas Hospital Research Centre, McGill University, Montréal, Québec, Canada}
\affil[3]{The Hub for Neuroengineering Solutions, University of Lethbridge, Lethbridge, Alberta, Canada}
\affil[4]{Canadian Centre for Behavioural Neuroscience, Department of Neuroscience, University of Lethbridge, Lethbridge, Alberta, Canada }

\corr{majid.mohajerani@mcgill.ca}{MHM}

\begin{document}


\maketitle

\begin{abstract}


The cattle industry has been a major contributor to the economy of many countries, including the US and Canada. The integration of Artificial Intelligence (AI) has revolutionized this sector, mirroring its transformative impact across all industries by enabling scalable and automated monitoring and intervention practices. AI has also introduced tools and methods that automate many tasks previously performed by human labor with the help of computer vision, including health inspections. Among these methods, pose estimation has a special place; pose estimation is the process of finding the position of joints in an image of animals. Analyzing the pose of animal subjects enables precise identification and tracking of the animal's movement and the movements of its body parts. By summarizing the video and imagery data into movement and joint location using pose estimation and then analyzing this information, we can address the scalability challenge in cattle management, focusing on health monitoring, behavioural phenotyping and welfare concerns. Our study reviews recent advancements in pose estimation methodologies, their applicability in improving the cattle industry, existing challenges, and gaps in this field. Furthermore, we propose an initiative \footnote{\href{https://www.opencattle.com/}{www.opencattle.com}} to enhance open science frameworks within this field of study by launching a platform designed to connect industry and academia. 
\end{abstract}

\section{Introduction}

The history of many human cultures is linked to the development of domesticated livestock, resulting in the formation of interdependent partnerships and economies between animals and people. Domesticated cattle have been central to some of these economies. Beginning with their first domestication between 11,000 and 9,000 B.C. \citep{lear2012our}, cattle husbandry has evolved from pastoral to include modern industrial farms. The contribution of the cattle sector to the economy of just Canada has been as much as 24 billion dollars on average from 2020 to 2022 \citep{canadabeef2023}. This substantial economic footprint underscores the urgent need for research to improve cattle welfare, productivity, and environmental sustainability.

Building on this foundation, efficiency emerges as the cornerstone of animal husbandry, particularly in beef production and broader cattle-based operations. With ever-increasing demand and limited resources, ranchers must maximize efficiency to thrive. Achieving higher efficiency involves enhancing output while maintaining or reducing inputs. This requires farmers to closely monitor their cattle, ensuring they are growing healthily, progressing at an acceptable rate, and being market-ready when the time comes.

Nonetheless, the complexity of modern farming introduces significant research challenges, particularly in animal health. Health monitoring focuses on disease detection \citep{poursaberi2009image}, effective reproductive management \citep{shahinfar2014prediction}, and improvements in feeding and overall animal welfare \citep{asher2009recent}. Previously, health monitoring mainly involved periodic visual inspections and walk-throughs by trained inspectors. However, in present-day farming practices, the costs associated with these methods can quickly exceed their benefits, especially as labor costs have risen over the past decades. Furthermore, traditional methods have consistently faced scalability issues.

New approaches include the integration of technology in the monitoring referred to as precision livestock farming (PLF) \citep{berckmans2017general}. Precision livestock farming is defined as “management of livestock by continuous automated real-time monitoring of production/reproduction, health and welfare of livestock and environmental impact” \citep{berckmans2017general}. Similar to many other technological advancements of the last decade, modern precision livestock farming is also tied with artificial intelligence (AI) \citep{garcia2020systematic}. A AI-based monitoring system is made from a few components, namely, a data collection module, a data processing module, and a user interface. In the data collection part, sensors or cameras are used to collect cattle data. Processing is where AI is used, where an algorithm is applied to the data to extract meaningful information. To make the extracted information accessible to a user who is not a computer expert, a user interface organizes the data to provide end-users with an overview. The data processing module is the core; it is where AI models are primarily utilized to perform their functions.

Among data collection schemes, computer vision has a special place. Computer vision is considerably inexpensive, as it allows a large number of animals to be monitored with one or more cameras. It is non-invasive, as it does not disturb subjects and, therefore, does not induce stress on livestock \citep{scoley2019using}. Computer vision also aligns with traditional observation-based techniques, and thus all the domain information used by experts in assessing health of cattle by observing them can be integrated into the monitoring systems.

\begin{figure}[htb!]
\centering
\includegraphics[width=\linewidth]{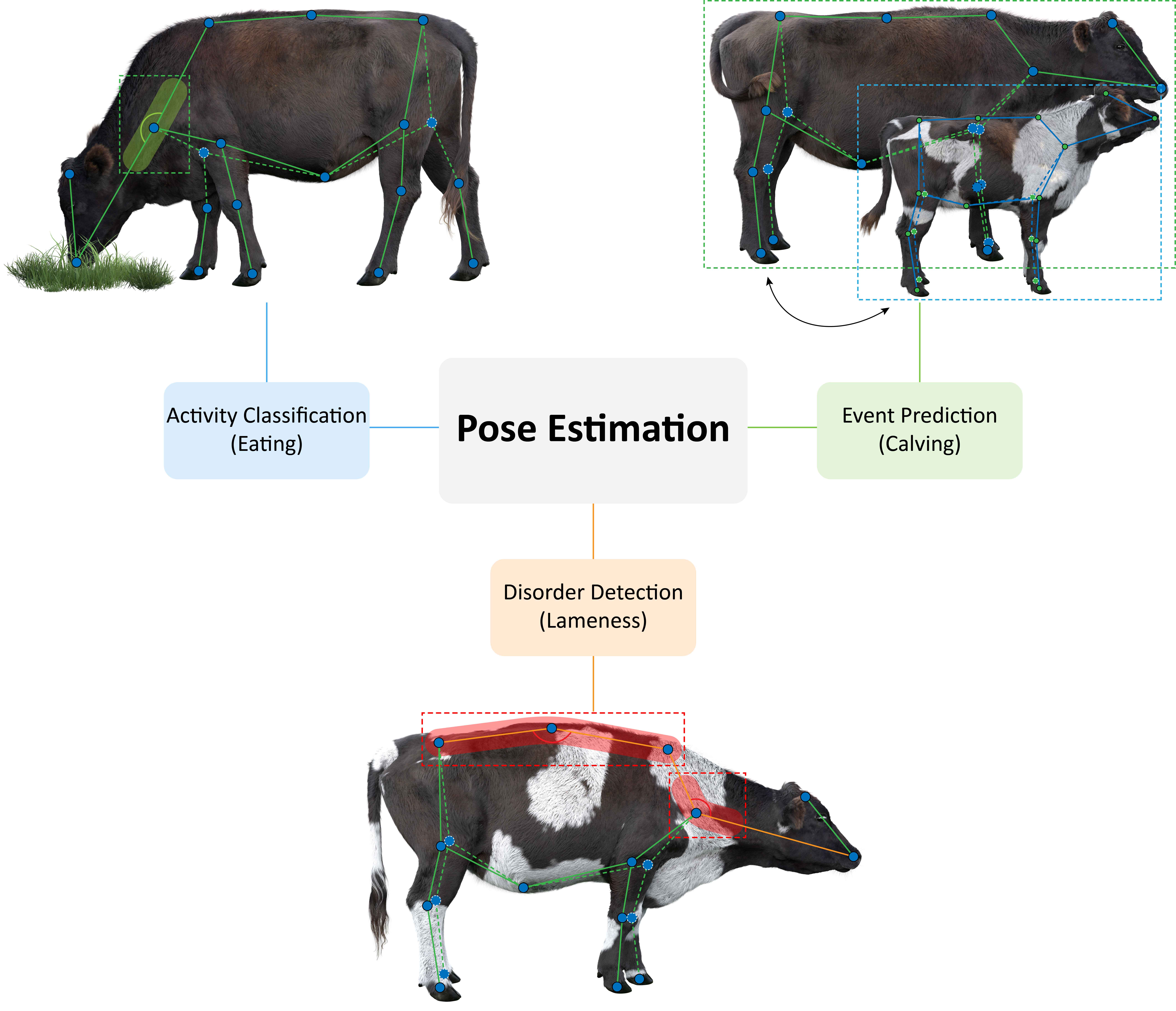}
\caption{Pose estimation can provide information regarding animals' behaviour (e.g., eating vs walking), specific events (e.g., calving), genetics (e.g., breed), animal age, and health conditions. \textbf{Top Left:} Eating behaviour, an activity that can be found in data using posture, and its variations through time. \textbf{Top Right:} Calving event, approximation of which can be estimated using posture changes through time \citep{speroni2018increasing}. \textbf{Bottom:} Lameness, detected by observing the bovine back arc \citep{sprecher1997lameness,scott1989changes}.}
\label{fig:applications}
\end{figure}

As part of a computer vision toolkit, pose estimation has a specific place in the development of a health monitoring system. Pose estimation enables precise identification and tracking of animal movement and the movement of animal body parts. Pose estimation can identify animal joints in visual data and thus help in assessing health problems such as lameness \cite{barney2023deep}. Its contribution to the detection of health issues is further emphasized by enabling behaviour and activity recognition \cite{wei2023study}. This is therapeutically advantageous because when health problems are identified early, subjects can be isolated from other animals and treated. In short, pose estimation allows for the study and monitoring of cattle 24/7, identifies deviations or irregularities from normal behaviour, and can signal potential health problems. Figure \ref{fig:applications} shows some pose estimation applications described in published research.

Given the significance of research in enhancing efficiency in animal husbandry and the crucial role of pose estimation, this work goes through the details of model development for cattle pose estimation (CPE). Our aim is to identify current applications, gaps, and trends in pose estimation, focusing on the challenges and potential directions for future research to address them. We have surveyed existing academic research papers and compiled a summary in the form of a table in the appendix. Our effort is to bridge the gap between cattle industry end-users, animal science researchers and AI experts in this field by enhancing accessibility for all stakeholders. Additionally, we address existing gaps and explore how both sides can collaborate to resolve them.

To increase resource accessibility, we aim to develop and introduce an initiative that bridges industry and academia in developing computer vision-based health monitoring applications for cattle. Toward this end, we introduce the \href{https://www.opencattle.com/}{Open Cattle}; a hub for open science enthusiasts working on cattle monitoring with AI. This platform also aims to improve the availability of resources and data for researchers and bring solutions one step closer to ranchers. Throughout the remainder of this text, we explore the objectives of this initiative and its potential contributions to addressing current challenges.

\section{Development Process for Pose Estimation Methods}

Deep learning, a branch of AI, has changed the course of computer vision in the past decade because it is the basis of almost all current state-of-the-art methods in computer vision \citep{chai2021deep}. The present work focuses on the applicability of deep learning models to behavioural phenotyping. Here, a model can be seen as a set of computer-based mathematical operations that, when applied to an instance of data, produces the desired output (which is the location of joints in our case). These models require a round of training to become tailored to the task at hand; training is the process of showing the model samples of data with desired output.

The collection of the training samples provides the training data. In computer vision methods, as the name suggests, the training set consists of visual data, images or videos. Collecting training data is the first step in method development. Training data also require ground truth, which consists of desired outputs. In our case, for each instance of image, the output is set of joint locations in pixel space of that image.

After assembling a training dataset and training the model, it must be evaluated to assess its parameters and structures and ensure that it will perform the task as expected by the experimenter. The development and evaluation process requires knowledge about the specific components of the model and the evaluation metrics. After evaluation, the model is ready to be integrated into a product. In the following section, we will review the details of each of these steps.

\subsection{Data, the fuel of proper model training}

With the application of AI in almost every aspect of human life, data, "the new oil of the digital economy" \citep{toonders2014data}, has become a valuable commodity for training deep learning models. "Data" in computer vision broadly embodies the collection of images used to train a model. Data can comprise a set of faces if the model is intended to be applied to face recognition, examples of speech if it is to be applied to speech recognition, and pictures of livestock if the model is to be applied to animal husbandry. There are two ways of obtaining data. One way is to collect thousands of samples manually, which is a time consuming and expensive process. Another way to obtain data is to create an artificial dataset in silico, which is intended to represent real-world scenarios.

The efficacy of machine learning and deep learning models depends on the availability and quality of the data used to train them \citep{luca2022impact}. AI models are described as data-hungry because the more data they have, the more accurate they become \citep{aggarwal2018neural}. Thus, a dataset containing large sample sizes with high diversity is necessary to train accurate and generalizable models.

To understand the importance of sample size in training a model for higher quality of results, one can look at the evolution of benchmarks in computer science. Take ImageNet, one of the most famous of such datasets which is used to train computer vision models for image classification, for example; this dataset comprises more than 1 million pictures \citep{deng2009imagenet}.

\begin{figure}[htb!]
\centering
\includegraphics[width=\linewidth]{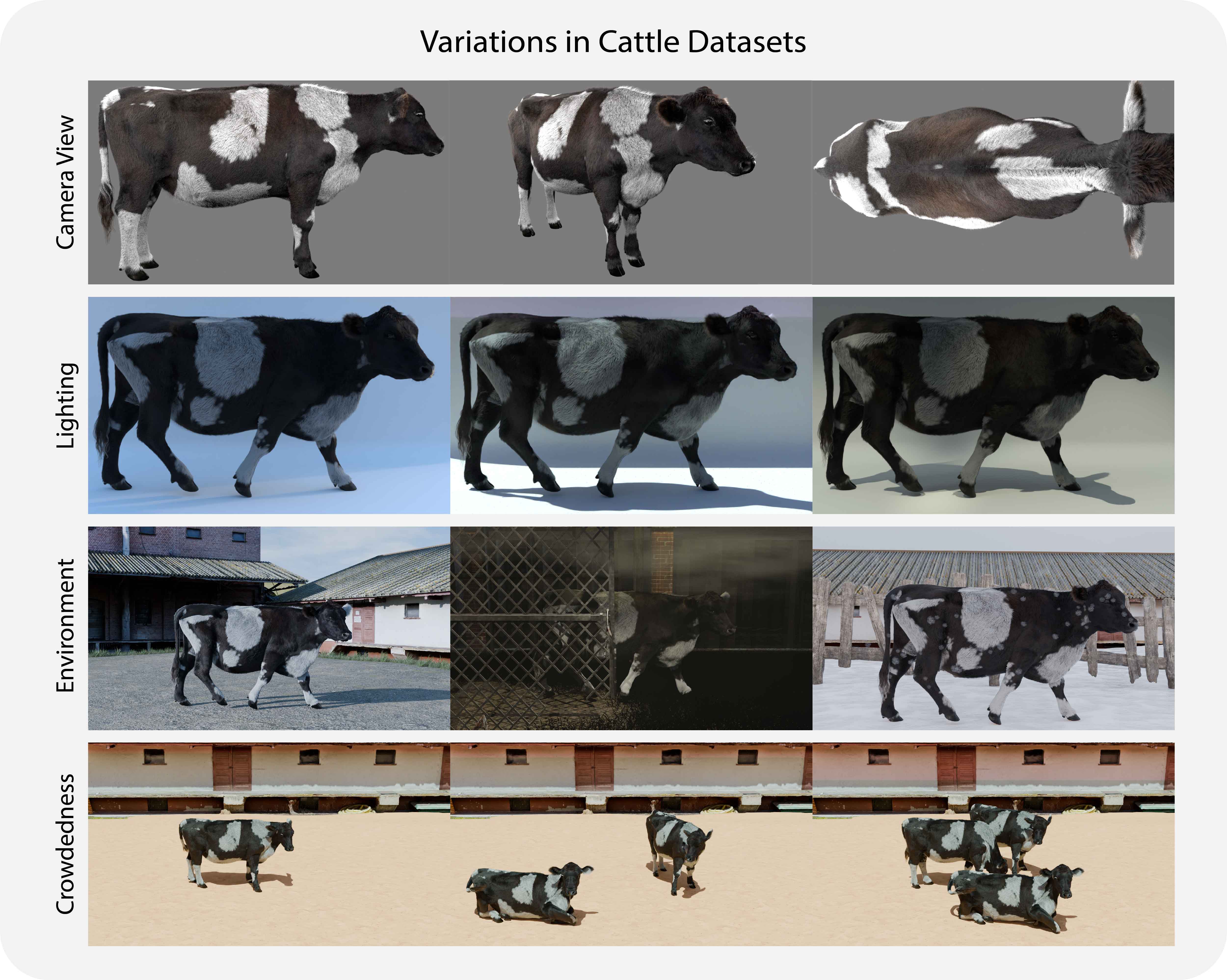}
\caption{Many parameters can cause a change in distribution of data for cattle behavioural phenotyping, some of which are shown in this figure. \textbf{Camera View:} By changing the view of camera, not only the data would change, but also some keypoints might become invisible. \textbf{Lighting:} The time of day and the lighting of the environment have a visible effect on data. \textbf{Environment:} Occlusions to view, or extreme weather conditions can come with different environments, and thus change characteristics of data. \textbf{Crowdedness:} Another parameter that introduces occlusions to the view and shifts to distribution is subjects overlapping in the captured image, courtesy of crowdedness in data.}
\label{fig:data}
\end{figure}

But the number of samples in the dataset is not everything. During inference, machine learning models tend to work best when applied to similar data to their training set. In other words, they work best when trained and tested on independently and identically distributed (i.i.d.) datasets \citep{goodfellow2016deep}. In cattle monitoring, there are several factors that can cause a difference in the distribution and instincts of the data. We have illustrated some major factors in Figure \ref{fig:data}, among many others. Moreover, factors such as type of cattle (which can be beef or dairy) implicitly enforce a change in many of the mentioned factors (as they have different habitat environments), thus changing the distribution of data.

When creating a model to be used in a commercial product, it is not feasible to go to each farm for data collection. The more interesting approach is to train the model once, but in a way that it can be applied everywhere. So, instead of matching the training distribution with the test one, researchers can expand the training distribution to ensure that it includes samples of the test distribution as well. In those terms, the final model can be considered more general; as it can be applied in almost everywhere.

Here, comprehensiveness is an important factor that should be considered when collecting data. Deep learning models are prone to overfitting to data and its characteristics \citep{ying2019overview}. This prevents the generalizability of the model and becomes troublesome when the dataset consists mainly of instances with high similarity. A good dataset captures different aspects and variations of real-world events, and thus, can be described as comprehensive \citep{gong2023survey}. Comprehensiveness can be seen in many details: having different subjects, collecting data from different angles, in different weather conditions, or even in different parts of estrous cycles.

In another attempt to overcome challenges of generalizability, it is common practice to combine a number of datasets to induce diversity and increase the size of the dataset. However, to combine datasets, they need to have similar ground truths, which here are the keypoints. A keypoint is the location on an animal or limb that a pose estimation method tries to find. Researchers annotate different keypoints; for example, in labeling limbs, most methods annotate hooves and knees, making a minimum of eight keypoints. Other methods also label the upper leg or the carpal \citep{li2019deep}. Labeling the head and back of the cow is even more variable. Some methods assign one point to the head \citep{fan2023bottom}, others might label the ears \citep{li2022fusion}, nose \citep{barney2023deep}, or mouth \citep{li2022basic}. When labeling the back arc of the cow; some methods use the shoulder, spine, and tailhead \citep{liu2020cow} others label two of these \citep{fan2023bottom}, or even one \citep{li2022basic}. The choice of a labeling scheme is mainly dependent on the joints that can be seen and the purpose of the study. However, a unified labeling paradigm can speed up research easily, as it allows for dataset combination without a hassle.

\begin{featurebox}

\caption{Public datasets (of animals) that contain instances of cattle}
\label{box:data}
\begin{multicols}{2}

While it is common practice to collect dataset according to the need of underlying task in this body of literature, there are some publicly available datasets which can be helpful to researchers for pre-training their models. Moreover, having access to other datasets enables researchers to compare their method and show superiority of their work easier. This also highlights the need for public benchmarks on CPE, to simplify the model evaluation process.

Although there is arguably no benchmark specifically and merely created for CPE, there are some well-known datasets on animals, which also contain images from cows as well. These datasets might be suitable for model evaluation and comparison, but mainly they provide researchers with great data for model pre-training and transfer learning \citep{weiss2016survey}. Moreover, there are a number of publicly available pre-trained models on these datasets, details of which can be found in Box \ref{box:code}.


\begin{adjustbox}{width=.5\textwidth,center}
\begin{tabular}{|c|c|c|}
\hline
\textbf{Dataset} & \makecell{\textbf{No. of Cattle}\\\textbf{Instances}}	& \textbf{Description}\\
\hline
\makecell{AP-10K\\\citep{yu2021ap}} & \makecell{323 instances\\of cows in\\200 images} & \makecell{Created with the objective of\\introducing a new benchmark for animal\\pose estimation in the wild, this dataset\\contains more that 10,000 images\\from 54 types of animals (13028\\total annotated instances)}\\
\hline
\makecell{APT-36K\\ \citep{yang2022apt}} & \makecell{47 video\\clips, 2400\\total frames} & \makecell{An extension to AP-10K dataset,\\this dataset additionally focuses\\on tracking as well, thus video\\clips where recorded from subjects}\\
\hline
\makecell{BADJA\\ \citep{biggs2018creatures}} & \makecell{104 images\\of same cow} & \makecell{This dataset has added joint annotation\\to DAVIS \citep{pont20172017} video\\segmentation dataset, and other videos\\for animal pose estimation}\\
\hline
\makecell{Animal-Pose\\ \citep{cao2019cross}} & \makecell{842 instances\\in 534 images} & \makecell{This animal pose estimation dataset\\contains 4000+ images, having more\\than 6000 instances of 5 types of\\animal: dog, cat, cow, horse, sheep} \\
\hline
\makecell{Poselets \\ \citep{bourdev2011poselets}} & \makecell{642 instances\\in 334 images} & \makecell{This dataset have added keypoints\\and foreground annotation to PASCAL\\VOC 2011 \citep{pascalvoc2011}\\dataset.} \\
\hline
\makecell{Animal3D \\ \citep{xu2023animal3d}} & \makecell{275 Ox and\\43 Water Buffalo} & \makecell{Aiming to create a diverse and\\comprehensive dataset for 3d animal\\pose estimation, this dataset is a\\premier in its scope}\\
\hline
\end{tabular}
\end{adjustbox}

\end{multicols}
\vspace{0.5pt}
\end{featurebox}

\subsubsection{How to access all the needed data?}

One of the main goals of the Open Cattle Hub is to gather datasets in this field of study to highlight gaps and scenarios where there is not much data available. We have already realized that datasets from crowded environments, such as feedlots, are lacking, and we have taken initiative to solve this issue; as a milestone of Hub, we aim to publish our dataset on crowded pose estimation in cattle (samples of which are available in Hub at the time of writing this).

However, in addition to identifying gaps, the main goal of the Hub is to connect academia and industry. With identified gaps, ranchers can easily provide researchers with the required data. Moreover, by testing the models in real-life, ranchers can provide their feedback and expectations directly to a group of researchers, and make sure that the products are going to be addressing the challenges they have.

Moreover, to open doors of open science, unification in the language of communication is necessary. To this end, Hub is taking some actions: 
\begin{itemize}
    \item \textbf{Unified Labeling Scheme:} As shown before, different labeling can cause a delay in adapting other datasets, as researchers need to re-annotate data to match their ground truth. To address this, we have chosen the labeling used in Animal-Pose \citep{cao2019cross}. This was due to the comprehensiveness of this labeling and its adaptation in one of the datasets with the highest instances of CPE.
    \item \textbf{Tools to Increase Accessibility:} After labeling, the ground truth is saved on disk; having a unified saving format can greatly contribute to increasing the pace of working with new data. We are adding tools to transform different file saving formats into the COCO dataset \citep{lin2014microsoft}, given its acceptance in computer science research. Notably, we have listed a number of tools that enable viewing data, and annotating them on COCO based formats.
    \item \textbf{Unified Benchmark:} Benchmarks help researchers develop better models by providing a platform of comparison that enables contributors to build on top of the prior work and create better results. We have a number of benchmarks on animal pose estimation (see Box \ref{box:data}) or human pose estimation \citep{zheng2023deep}, but the need for one specifically built for cattle pose is felt here. To this end, by combining available datasets and adding crowded scenarios to them, Hub aims to create a benchmark for pose estimation as well.
\end{itemize}

\subsubsection{Synthetic Data, a Workaround Solution}

Another promising avenue of research is the generation of synthetic data and data augmentation. With synthetic data, researchers can expand their datasets, allowing them to train larger models \citep{de2022next}. Synthetic data can also be used to create instances of data from scenarios where data collection is costly, time-consuming, or is not possible on a large scale. Synthetic data generation in the field of animal pose estimation has seen some advancements in the past couple of years \citep{jiang2022prior}, and CPE can benefit from it.

\subsection{Training a deep learning model}

After collecting the data and labeling it, the next step would be to create a deep learning based model, and train it. These models have a structure that needs to be designed for the task at hand \cite{goodfellow2016deep}, then there are a number of wights (variables) in them that need to be optimized, in 
a process called model training. Designing a model structure from scratch is not usually a common choice here. Creating and designing models are usually left to computer scientists; rather than that, it is usually common to take a model structure, and tailor-make its features for the task at hand. Thus, in the following, we have gathered two pieces of information necessary for this process: A) an overview of different components in models and their role and B) state-of-the-art and well-established models.

\subsubsection{Model Component}

Models have different components and stages, each with a different objective. Here, we discuss the main segments of a model.

\paragraph{Pre-Processing}

The main objective of pre-processing is to prepare data for model training. A pre-processing pipeline can also be considered separate from the model, as it operates independently before the model execution. The absence of a proper pre-processing module can hinder the optimal performance of the models \citep{chaki2018beginner}. Actions such as normalizing images \citep{pal2016preprocessing}, removing redundant information \citep{vu2018non}, and enhancing contrast \citep{perumal2018preprocessing} all are tasks that fall under the pre-processing.

\paragraph{Backbone}

Following pre-processing, the next segment where data goes through is the backbone. This part is usually a neural network that performs the feature extraction segment of the model. Utilizing pre-trained neural networks as backbones is common, often with fine-tuning for a specific task. Even when trained solely on ImageNet \citep{deng2009imagenet}, these networks perform well in extracting information and serve as a solid foundation for the model \citep{cao2017realtime}.

\paragraph{Neck}

The neck of a pose estimation model has features extracted from the backbone for further processing to generate finer features before using them to create output.  Although an optional component, the neck plays a pivotal role in tailoring the model to particular dataset needs. There are several model necks, the most prominent ones being the pyramid feature network \citep{lin2017feature} and the path aggregation network \citep{liu2018path}. Although exploring the details of these models is out of the scope of our work, it can be mentioned that both are especially advantageous when datasets have subjects of varying sizes, enabling the model to efficiently extract information on different scales \citep{chen2021empirical,yu2023multiple}.

\paragraph{Head}

The head segment of a pose estimation network is responsible for producing outputs. But implicitly, since it enforces how the loss function works (the criteria for optimizing model weights), it also impacts the training and learning process of the network. There are two common types of heads. Regression-based heads predict precise coordinates via continuous outputs that represent $x$ and $y$ of each joint. Heatmap-based heads generate a heatmap sized to the image, highlighting regions of interest. The latter is more common in DL-based models, as they allow a more natural and superior gradient flow for model optimization. To generate labels for heatmap-based models, one approach is to put 2D Gaussians, centered on the ground truth for that joint (and for the standard deviation, fixed values, such as 1 pixel \citep{li2019deep} or flexible values based on joint size or subject scale).

In both models, one output head (heatmap or regressor) is required per joint. Note that predicting the absence of a joint is also more natural and straightforward in heatmap-based approaches.

\subsubsection{Handling Multi-instances}

A pose estimator model is well suited to record information on joints and body parts in one subject. When there is more than one subject present in the image, other measures must be taken. There are two main approaches that deal with multiple images \citep{zheng2023deep}: 
\begin{enumerate}
    \item \textbf{Top-Down Methods:} A simple solution to enable methods to train properly when multiple subjects are presented, is to individuate the subjects, and then, try to detect joints on each subject separately \citep{ning2018top}. 
    \item \textbf{Bottom-up Methods:} Bottom-up methods first try to detect joints where possible and then merge them to form a composite pose \citep{cao2017realtime}. 
\end{enumerate}

Note that when there is merely one subject presented, both types can be used in their natural form, but it is common to remove some complexities of the model that deal with multi-subject nature of data (e.g., remove the subject detection stage in top-down models).

\subsubsection{Segmenting Body Parts}

Another group of methods that seek to find structure of the skeleton is body part segmentation \citep{antink2020fast}. In contrast to pose estimation methods, these methods look for a bounding box for each body part \citep{jiang2019flyolov3}; and in some more complicated case, they work on finding exact pixels in which a body part is presented \citep{antink2020fast}. These methods also have structures and outcomes similar to pose estimation methods; therefore, we have also reviewed some of their applications in the Appendix table.

The heads of these methods can also have both heat map and regression shape; the only difference is that in the regression form, it outputs the bounding box surrounding a specific body part.

\subsubsection{State of the art methods}

Researchers have presented various approaches in model structure designs for pose estimation. This section of the paper is devoted to discussing the state-of-the-art design paradigm and models.

First, a common practice is to follow the selection of majority; here, the HRNet \citep{wang2020deep} and DeepLabCut (DLC) \citep{mathis2018deeplabcut} networks stand out. These two networks appear more than any other selection in research in this field (see the Appendix Table). HRNet operates as a multistage network, featuring several parallel paths at each stage. Each path processes images at a different scale, which helps the network adapt to and find objects of various sizes. These networks are still considered as the baseline benchmark in many HPE works due to their simple intuition and high performance. DLC networks, while typically employing a straightforward ResNet \citep{he2016deep} as their feature extraction backbone, have demonstrated remarkable performance in a variety of animal pose estimation tasks, earning them recognition among researchers in the field \citep{mathis2020primer}.

Another approach in developing models for pose estimation is to adapt an object detection framework and change it to be suitable for pose estimation. CPE models address similar challenges encountered in object detection, such as managing varying sizes, producing detailed output, and recognizing multiple instances within a single image. Examples of object detection networks that have been adapted in CPE are the YOLO family \citep{hua2023effective,jiang2019flyolov3}, Mask RCNN \citep{barney2023deep,wang2023op}, and Faster RCNN \citep{gardenier2018object}. They are also more common in body part segmentation tasks, given their inherent suitability for it without necessitating modifications to the network design.


Part affinity fields (PAFs) \citep{cao2017realtime} represent another area of interest in pose estimation, and networks based on this concept are widely acknowledged in the literature. PAFs provide a mapping for the location and orientation of body parts through 2D vectors. In bottom-up approaches, networks equipped with a keypoint detection module can leverage PAFs to effectively combine keypoints into a complete pose, especially in densely populated scenes \citep{cao2017realtime}. The application of graph-matching algorithms \citep{bunke2000graph} is a strategy used in this process.

Lastly, there are a number of more recent methods applied in the human pose estimation, which have not yet found their way into CPE. The vision transformers \citep{dosovitskiy2020image} has markedly transformed the expectations and capabilities of models in computer vision \citep{khan2022transformers}, with VitPose \citep{xu2022vitpose} and Swin-Pose \citep{xiong2022swin} being examples of the potential of these networks in pose estimation. These methods introduce a direction for future research, and integration of them in CPE can improve performance.


\subsubsection{Selecting Best Model}

The choice of method is mainly dependent on the objective of the study; HRNet and DLC can be considered a safe choice, while transformer-based models present room for better results. Bottom-up models, on the other hand, can help in crowded scenarios, where the bounding box of a subject contains parts of other subjects. Also, another important aspect is implementation and pre-training the model. Pre-training is the process of training model on similar datasets (for example, here, training the model on a complete animal pose estimation dataset), to have a better starting point for training on main data. Box \ref{box:code} provides some information about the libraries available for implementation.

\begin{featurebox}
\caption{Libraries and Coding}
\label{box:code}
\begin{multicols}{2}
    When we look at the implementation of CPE methods, there are means available for anyone from any level of coding skills. At the lowest level, these methods can be seen as deep neural networks and thus can be implemented with any deep learning library (e.g., PyTorch \citep{paszke2017automatic} and TensorFlow \citep{abadi2015tensorflow}). But there are some libraries specifically made for pose estimation. MMPose \citep{contributors2020openmmlab}, which is one of the libraries of the OpenMMLab family, is a standout here. This library has implementation of many famous network structures, such as HRNet, YOLO-Pose \citep{maji2022yolo}, RTMPose \citep{cai2020learning}, and many more \footnote{\href{https://mmpose.readthedocs.io/en/latest/model_zoo_papers/algorithms.html}{https://mmpose.readthedocs.io/en/latest/model\_zoo\_papers/algorithms.html}}. In addition to algorithms, they also have a zoo from different datasets, and more importantly, a number of pre-trained models on various datasets, such as well-known animal pose estimation datasets, such as AP-10k and AnimalPose \footnote{\href{https://mmpose.readthedocs.io/en/latest/model_zoo/animal_2d_keypoint.html}{https://mmpose.readthedocs.io/en/latest/model\_zoo/animal\_2d\_keypoint.html}}.
    Despite its resources, working with MMPose requires some knowledge of programming, and can be tedious for someone with less coding experience. When switching to higher level tools, DLC \citep{mathis2018deeplabcut} is definitly the most used library. This library has tools for annotation, and allows the user to do the whole process of model training by a simple graphical interface, without doing any coding. However, DLC is very limited in terms of available models, and can not compete with MMPose on that front.
    \end{multicols}
\end{featurebox}

\subsection{Model Evaluation}

Once a model has been created and tailored to the requirements of the data set, its performance needs to be evaluated. This evaluation allows for selecting hyperparameters, network designs, and also making sure that the model would maintain its performance when applied in product.

Two main considerations in conducting an evaluation are metrics to use and the evaluation process. The metrics provide statistical insight into the performance of the model. Although some metrics, such as accuracy, offer straightforward evaluations without necessitating intuition, others may be more challenging to interpret. Box \ref{box:metrics} goes through a number of metrics used and how to calculate them, but most libraries have already developed codes for metric calculations, and researchers can use them in a plug-and-play manner. Next, a discussion is presented on designing a reliable evaluation process.

\begin{featurebox}
\caption{Evaluation Metrics}
\label{box:metrics}
\begin{multicols}{2}

The metrics provide statistical insight into the performance of the model. Although some metrics, such as accuracy, offer straightforward evaluations without necessitating intuition, others may be more challenging to interpret. In this box, we have outlined several commonly used metrics in CPE research.

\begin{itemize}
    \item \textbf{PCK:} The Percentage of Correct Keypoints (PCK), as introduced by \citep{yang2012articulated}, measures the precision of detected keypoints by determining the proportion that correctly falls within a predefined threshold around the ground truth. The selection of this threshold is significantly influenced by the size of the subject in the image, which in turn depends on the distance from the camera. Typically, the length of the head of the cattle is used as a normalizing factor, resulting in the metric being called PCKh. Furthermore, the choice of an acceptable normalized threshold varies in different studies. However, a threshold of 0.05 is often considered indicative of high precision \citep{jiang2022animal}, with a range of 0.05 to 0.3 typical in various works \citep{jiang2022animal}. The final formula to calculate PCKh is as follows: \newline \begin{equation}
        PCK = \frac{\sum_{n=1}^N\sum_{i=1}^M \delta (\frac{d_{n_i}}{h_n} < \alpha) * v_{n_i}}{\sum_{n=1}^N\sum_{i=1}^M v_{n_i}}
    \end{equation}
    where $N$ is the number of instances of cattle in the test dataset, $M$ is the number of keypoints, $\alpha$ is the threshold, $v_{n_i}$ is the visibility of the keypoint $i$ in instance $n$ ($1$ if visible, $0$ otherwise), $h_n$ is the head length of instance $n$, $d_{n_i}$ is the distance between prediction and grand truth, and finally the $\delta$ function is $1$ if the condition is met, $0$ otherwise. \newline
    Also, some works change the nature of visibility parameter; as they expect the model to predict even in invisible scenarios. In that case, the formula should also be modified accordingly. 
    \item \textbf{AP:} If we consider PCK to be accuracy in pose estimation world, we can similarly create metrics for precision and recall \citep{bishop2006pattern} as well. Average precision (AP) refers to the area under the precision-recall curve or average of precision for a class of objects at different thresholds \citep{padilla2020survey}. If we average these values over all classes in dataset, the mean AP (mAP) is obtained. Average recall (AR) is more common in the literature on object detection, where the recall is averaged over different numbers of accepted detections (sorted by confidence) \citep{padilla2020survey}.
    \item \textbf{RMSE:} Mean squared error (MSE) is a well-known indicator of how well a model has predicted continues values. As indicated by the name, MSE is the mean of error (distance of prediction and ground truth) squared. Since the value of MSE is difficult to interpret given the exponential component, it is common to report and use squared root of MSE, or RMSE. Finally, RMSE is calculated using the following equation.
    \begin{equation}
        RMSE = \sqrt{\sum_{n=1}^N \sum_{i=1}^M d_{n_i}^2}
    \end{equation}
    Here, the definition of parameters is the same as in PCK. The main shortcoming of this metric is the fact that it is solely dependent on distance, thus one significantly wrong prediction can show lower performance, than a completely wrong prediction, but each keypoint by a small amount.
    \item \textbf{OKS:} Object keypoint similarity (OKS) is a metric known for its use in the evaluation of the performance of the model in the COCO \citep{lin2014microsoft} keypoint challenge. This metric is calculated as follows: 
    \begin{equation}
        KS_{n_i} = \exp (\frac{-d_{n_i}}{2s_n^2k_i^2})
    \end{equation} 
    \begin{equation}
        OKS_n = \frac{\sum_{i=1}^M KS_{n_i} \delta(v_{n_i} > 0)}{\sum_{i=1}^M \delta(v_{n_i} > 0)}
    \end{equation}
    Where $s_n$ is the scale of instance $n$ (in the COCO dataset, $s_n^2$ is equal to the area of the object's segment), and $k_i$ is the keypoint constant. Visibility in COCO dataset is can take values of $0$, $1$, or $2$. $0$ indicating unlabeled keypoints, $1$ is labeled but not visible, and $2$ is labeled and visible. The definition of other parameters is the same as in PCK.
\end{itemize}
\end{multicols}
\end{featurebox}

\subsubsection{Reliable Model Evaluation}

Deep learning (DL) models are notorious for a well-known weakness, overfitting to training data \citep{ying2019overview}. Without a proper test design, reported metrics may also suffer from similar phenomenon. In its simplest form, the training and test data must be different. However, evaluating the generalizability of the model can pose challenges, as test data might closely resemble training data rather than situations that can occur when the model is deployed \citep{zhou2022domain}.

As discussed above, ensuring model generalizability requires dataset diversity. Various factors that affect the data, such as the color or size of a cow, the temperature, and the time of day, should be considered. In addition, data instances should be collected in different settings to smooth the contribution of such factors. In test design, to ensure robust model evaluation, creating a comprehensive dataset is important. After creating a comprehensive, to validate model generalizability, one method involves manually splitting the train and test data in a manner where a specific value for a factor (such as cattle color) is not present within the training data. This is similar to the procedure used in the domain generalization literature \citep{zhou2022domain}. This type of meticulous testing is challenging and necessitates that a researcher account for domain generalization in their method development, which is not commonly done.

In the few samples that researchers have accounted for domain generalization, the most commonly used practice involves splitting data based on visual factors of the subject \citep{russello2022t}; while other means of splitting, such as weather variation \citep{wu2021using} are also applied.

In a final note, in research, result re-production is essential; in computer vision, this can be achieved by following a number of steps and sharing codes, data, and also trained models. While we understand that it is not possible for every research paper to follow these steps, in cases that is possible, it can greatly help those who do follow up research. AAAI also has a reproducibility checklist that can greatly help researchers in this field \footnote{\href{https://aaai.org/conference/aaai/aaai-23/reproducibility-checklist/}{https://aaai.org/conference/aaai/aaai-23/reproducibility-checklist/}}.

\section{Processing the Pose Estimation Outputs}

Up until now, we have discussed the process behind making a method for extracting the posture. As shown in Figure \ref{fig:applications}, pose has a number of applications in cattle monitoring. We shortly discuss how pose is used in applications, and then move to commercial products.

First and foremost, as discussed above, the pose can be used to detect disease. Lameness stands out here; it is a symptom of a number of diseases, some of which can be easily treated if detected early \citep{milner1997effects}. Furthermore, as recent studies show, lameness itself is responsible for 16-40\% of health-related problems in feedlots \citep{fortysixty}. There are several posture-based features that can be used to detect lameness \citep{bradtmueller2023applications}, such as head position, height of the spine markers, length of the spine, length of the thoracic region, and length of the lumbar region.

Combining spatial features with temporal ones opens the doors to extracting many other information, such as body movement scoring \citep{higaki2024leveraging}, walking speed, and analysis of individual limb movement \citep{franco2016investigation}. Overall, \citep{bradtmueller2023applications} has done a great job reviewing papers in this field and can help readers find methods for post-processing posture to detect lameness.

Classification of activity and action can easily enable the analysis of behavioural patterns. Behavioural studies also, in turn, enable analysis on welfare, health and productivity \citep{chen2021behaviour}, and can also help in breeding selection and predicting growth performance \citep{richeson2018using}. All of these allow systems to provide strategies for improving efficiency of farm, whether its beef or dairy cattle. Lastly, detecting abnormalities and events of interest (e.g., calving \citep{speroni2018increasing}).

\section{Commercial Products}

Considering the economic importance of animal husbandry and particularly the role of cattle within this sector, numerous commercial applications for monitoring and efficiency increasing have already been developed. A diverse array of solutions exist here, ranging from basic applications for record keeping to advanced systems that monitor the behaviour patterns of individual cattle and aim to address health related issues.

Many of these products currently work with hardware-based sensors, such as wearable ones. These types of data collection sensors usually have a high maintenance cost; can easily be damaged and need repair or replacement. As a result, a computer vision-based solution can help reduce costs significantly. Additionally, wearable sensors are not suitable for all environments. In Alberta, Canada, for example, the temperature can drop as low as -40 degrees Celsius; in order to create a working device for these environments, costs of material (for example, a battery that survives long enough) exceed the benefits they provide, and thus many farmers would rather stick to old-fashioned human-centered monitoring.

Nevertheless, we have checked products with two objectives in mind; firstly, where can pose estimation help create systems that can improve the efficiency of existing systems, and secondly, where have they been already implemented. We have summarized a number of example products in Table \ref{tab:cp}.

\begin{table}[htb!]
\caption{Examples of commercial products in cattle monitoring.\label{tab:cp}}
\centering
\resizebox{0.9\linewidth}{!}{
\begin{tabular}{ccc}
\hline
\textbf{Product} & \textbf{Application}	& \textbf{Sensor Type}\\
\hline
\makecell{\href{https://www.afimilk.com/cow-monitoring/}{Afimilk}} & \makecell{Health monitoring, pregnancy rate improvement\\suggestions, behaviour monitoring, eating\\analysis, milk quality improvement strategies} & \makecell{Wearable (Colar,\\Ankle Band)}\\
\hline
\makecell{\href{https://agtech.folio3.com/cattle-counting-software/}{Folio3 - cattle counting}} & \makecell{Counting} & \makecell{Vision (with drone or\\installed cameras)} \\
\hline
\makecell{\href{https://moonsyst.com/bolus}{Moonsyst}} & \makecell{Health monitoring (with temp, ph, etc), calving\\and eating monitoring, drinking behaviour} & \makecell{A device called bolus\\that is eaten and stays\\in reticulum of animal\\during lifetime}\\
\hline
\makecell{\href{https://www.cowmanager.com/}{Cow Manager}} & \makecell{Nutrition, health, ferility monitoring\\(through movement and temperature)} & \makecell{Ear sensor\\(similar to ear tag)}\\
\hline
\makecell{\href{https://herd.vision/}{Herdvision}} & \makecell{Body condition scoring, mobility scoring,\\monitoring system and mobile app} & \makecell{Camera} \\
\hline
\makecell{\href{https://www.onecup.ai/betsy}{BETSY}} & \makecell{Calving event notification, breeding feedbacks,\\disease detection in future, and connection\\to OneCup AI herd management application} & \makecell{Camera}\\
\hline
\makecell{\href{https://vytelle.com/vytelle-sense/}{Vytelle SENSE}} & \makecell{Measuring feeding intake, feeding behaviour,\\partial body weight, and growth} & \makecell{RFID ear tags,\\scales}\\
\hline
\makecell{\href{https://www.ever.ag/dairy/software-solutions/cainthus/}{Cainthus}} & \makecell{Monitors nutritional, behavioural, health,\\environmental activities, early\\flagging of illnesses} & \makecell{Camera} \\
\hline
\end{tabular}}
\end{table}

As demonstrated in the table, one specific step in health monitoring is to find drinking, eating, calving, and other behaviour patterns. Detection of these behavioural patterns can also happen in lower-level indicators, such as movement and mobility scoring as well. In the research, these tasks have been shown to be detected and categorized by pose estimation and action recognition models in the research, and this has been discussed earlier in this paper. Another interesting product in the table focuses on the estimate of cattle weight, a task that researchers have achieved using pose estimation \citep{liu2023feature}.

Pose estimation methods offer significant advantages beyond cost reduction. They provide information on the structural pose and complex behaviours of cattle, which are challenging to detect using other sensor data. For example, uneven weight distribution, indicative of lameness \citep{newcomer2016distribution}, can be more easily identified. Wearable sensors have unique benefits, such as measuring temperature, a metric not effectively captured by computer vision. While thermal cameras exist, they are not always suitable in open feedlots. Given these considerations, the potential for future products that utilize hybrid models that integrate both pose estimation and wearable sensors is promising.

The challenge of identification remains when we switch from wearable sensors to cameras. We need to identify subjects to continuously monitor them. Cows are often similar in color and size, making identification difficult. Some studies address the problem of similarity. These systems can use some clues, such as required ear tags \citep{smink2024computer}, or can use implicit visual cues, such as pose \citep{wang2023op}. These methods can also be used to improve counting applications, which is another group of commercial products. Notably, face identification can be done for cattle \citep{xu2022cattlefacenet}, but it is mainly impractical if subjects are far from the camera. Lifelong identification of cattle using computer vision without invasive indicators can also be seen as a challenge, and we might add it later on to the Hub.

Moving toward applications of computer vision in products that currently exist, we first have simple products that perform counting with drones. These products are useful for pastor-based animal husbandry. Next, we have products that perform higher level analysis, such as mobility scoring, and behavioural analysis. These products can be seen as proof that computer vision can help this field enormously.

It should be noted that pose estimation has also been integrated into products. BETSY, a product of OneCup AI, is one of the most promising products in this field. BETSY uses pose estimation and body segmentation and can perform cow face identification. They are working to add disease detection as well. Their product, while promising, faces the challenges of needing a well-structured research basis and proof of compatibility with various scenarios.

Nonetheless, we should never forget that industrialization of research and its practical applications can attract funding to the field, which in turn enables further research. Pose estimation has been recognized as a tool for behavioural analysis in neuroscience \citep{pereira2020quantifying}, and there is no reason for it not to have same place in industrial applications.

\section{Conclusion and Future Directions}

Open science democratizes knowledge, fostering innovation, and global collaboration; it is the future because shared understanding accelerates progress in every field. This concept is grounded in the belief that collective intelligence and open access to information can lead to faster and more equitable advancements across disciplines.

In the evolving landscape of cattle monitoring technologies, we stand at a pivotal juncture that beckons a shift from traditional monolithic solutions towards a more modular and collaborative approach. While comprehensive, the current generation of products encapsulates functionalities from A to Z in tightly sealed packages. Although effective in offering a one-stop solution, this approach inadvertently stifles innovation by limiting contributions to the enhancement of individual components within the system.

The essence of innovation lies in the nuanced improvement of each aspect of technology. Recognizing that these systems are conglomerates of smaller interdependent components, we can pave the way for a future where enhancement of individual elements directly contributes to the overall efficacy and sophistication of monitoring solutions.

Hub envisions a transformative direction for the research and development of cattle monitoring technologies. Our aim is to disentangle the existing all-in-one product paradigm and move toward an ecosystem where each component of monitoring technology is an open field for innovation. By establishing a unified language and interoperable standards between these components, we not only hope to facilitate seamless integration, but also democratize the process of innovation. This open ecosystem approach enables a broader spectrum of contributors, from individuals and startups to established corporations, to lend their expertise and insights to specific aspects of the technology. Whether it is advancing sensor accuracy, refining data analytics algorithms, or enhancing user interfaces, each contribution enriches the collective efforts.

Recognizing that the epicenter of research, particularly within AI-driven studies, often gravitates towards the data processing module, we have dedicated this work to delving into one of the field's cornerstone tools: pose estimation. This examination has illuminated its potential to unlock a spectrum of applications, simultaneously paving the way for the development of new products and the refinement of existing solutions in this domain. Moreover, the main technological goal of Hub is enabling enhanced research into tools specifically tailored for cattle pose estimation.

Lastly, opening the box of cattle monitoring technologies to modular contributions does not simply expand the horizon for product development. It embodies a profound shift toward inclusivity and collective progress. It ensures that advances in technology are not monopolized by those with the most resources but are instead driven by the best ideas, irrespective of their origin. Through Hub, we invite thinkers, innovators, and creators from all walks of life to join in this collaborative journey, enriching our shared mission with their unique perspectives and skills.

As we embark on this path, we not only foresee a future of more advanced and precise cattle monitoring tools, but also one where the spirit of collaboration and open innovation propels us towards unforeseen heights of achievement. The future of cattle monitoring is not just in the hands of a few but in the collective effort of many, each contributing to a larger and cohesive vision of technological excellence and sustainable agriculture.

\section{Acknowledgement}

This work was supported by fund provided by Alberta Innovate and Natural Sciences and Engineering Research Council of Canada (grant 390930) to Majid H. Mohajerani.

\bibliography{main}


\appendix

\pagebreak
\onecolumn

\begin{landscape}
    \tiny
    \newgeometry{left = 20pt, right =-100pt}
\begin{longtable}{|c|c|c|c|c|c|c|c|c|c|c|c|}
\caption{\centering Review of related works.}
    \label{tab1}\\
    \hline
    \textbf{Reference} & \textbf{Type} &  \textbf{Size}	 & \makecell{\textbf{No. of}\\ \textbf{Subjects}}	& \makecell{\textbf{Keypoints or Body Parts}}	& \textbf{Crowdedness} & \textbf{Data Acquisiton Protocol}	& \textbf{Category} & \textbf{Method Overview} & \textbf{Head} & \textbf{Application} & \textbf{Highlights}\\
    \hline
    \citep{li2019deep} &
    \makecell{Dairy\\and\\beef} &
    2134 images &
    \makecell{33 dairy /\\30 beef} &
    \makecell{16: Head, Neck, Spine,\\Coccyx, 12 on Legs\\(Roots, Knees, Hooves)} &
    \makecell{Single Subject\\(there might be\\unannotated subjects\\in background)} &
    \makecell{\textbf{Device}: iPhone 8 plus\\\textbf{VRS}: 1080p, 30fps, 8sec\\} &
    \makecell{Single\\Subject} &
    \makecell{
        A) Preproceesing by cropping and resizing\\
        B) Data augmentation to increase samples by\\
        rotation, horizontal flip, and brightness\\
        C) Body part detection using a \textbf{VGG-FCN}\\
        based on \textbf{VGG-16}\\
        D) Keypoint detection using a light \textbf{VGG-FCN}\\
        on output of previous step
    } &
    Heatmap &
    \makecell{Pose\\Estimation}&
    \makecell{Publicly Available Dataset,\\Comparison of 3 network\\structures for the task}\\
    \hline
    \citep{fan2023bottom} &
    \makecell{Dairy\\and\\beef} &
    \makecell{2432 images\\(with 3101\\instances\\in them)} &
    \makecell{Two datasets:\\1. Same as\\
    \citep{li2019deep}\\2. Not described} &
    \makecell{16: Head, Neck, Spine,\\Coccyx, 12 on Legs\\(Roots, Knees, Hooves)} &
    \makecell{Images with more\\than one subject,\\but no info on IOU} &
    Not applicable &
    Buttom-Up &
    \makecell{Four-stage \textbf{HRNet} with modifications:\\
    1. less populated stage 3\\
    2. Increased blocks in stage 3\\
    3. Attention at the end of stage 1\\
    4. Layers in stages 2-4 by a\\
    combination of depth-wise and\\
    normal convolutions changed\\
    5. Adaptive convolution in\\
    keypoints head}&
    Heatmap &
    \makecell{Pose\\Estimation} &
    \makecell{All subjects, even\\distant ones are labeled,\\Model is compared\\to others and superiority is\\shown}\\
    \hline
    \citep{liu2020cow} &
    \makecell{Dairy} &
    \makecell{219 10s\\videos\\1495 frames\\in total} &
    Not mentioned &
    \makecell{17: Nose, Head, Top of\\neck, Bottom of neck, Shoulder,\\Spine, Tailhead, Mid-thigh,\\Bottom of shoulder, 8 on\\Legs (Hooves, and another\\ point higher on Legs)} &
    \makecell{Multi Subject\\but IoU=0} &
    \makecell{\textbf{Device}: GoPro, DVR,\\high-quality IP camera\\\textbf{View}: Side\\\textbf{Env}: Walking Path} &
    Bottom-Up &
    \makecell{
    A) Keypoint detection with 2 \textbf{CNN}s:\\
    1. \textbf{DeepLabCut (DLC)} \citep{mathis2018deeplabcut}\\ with \textbf{ResNet} 2. Custom \textbf{CNN} \\
    B) 3-stage post-processing:\\
    1. Body part detection\\
    2. Distinguishing instances, missing\\
    point prediction, keypoint validation, and\\
    outlier removal 3. Temporal smoothing 
    } &
    Heatmap &
    \makecell{Pose\\Estimation} &
    \makecell{Model capacity in detection is\\also tested.}\\
    \hline
    \citep{gong2022multicow} &
    Dairy &
    \makecell{2200 images\\for detection,\\1800 cow\\instances\\for pose\\estimation} &
    10 &
    \makecell{16: Head, Neck, Spine,\\Coccyx, 12 on Legs\\(Roots, Knees, Hooves)} &
    \makecell{There are occlusion\\of different cow\\but IoU is\\not mentioned} &
    \makecell{\textbf{VRS}: 1080p, 25fps, 25min \\ \textbf{CP}: 4m high \\ \textbf{Env}: 10 cows, 35m*20m} &
    \makecell{Hybrid} &
    \makecell{
    A) Cow Detection using \textbf{YOLOv4}\\
    with \textbf{CSPDarknet53} B) Feature\\
    extraction using \textbf{VGG-19} C) Keypoint\\
    detection using a multi-stage network\\
    D) Pose estimation using graph-matching\\
    E) Pose classification using a\\
    \textbf{MLP}-like structure
    } &
    Heatmap &
    \makecell{Activity\\Classification\\(lying, standing,\\walking)} &
    \makecell{
        Performance metrics of the model\\
        was reported on sub-optimal\\
        conditions. Pose classification\\
        network reached high precision.
    }\\
    \hline
    \citep{wei2023study} &
    Dairy &
    \makecell{3900 images\\for pose\\estimation,\\1800 frame\\groups for pose\\classification} &
    \makecell{3 (with\\another 6 not\\labeled)} &
    \makecell{16: Head, Neck, Spine,\\Coccyx, 12 on Legs\\ (Roots, Knees, Hooves)} &
    Not mentioned &
    \makecell{\textbf{Device}: 5MP Camera \\ \textbf{CP}: 4m high \\ \textbf{Env}: 9 cows, 35m*20m} &
    Bottom-Up &
    \makecell{
    A) Keypoint detection using a\\
    multi-stage network B) Pose estimation\\
    using graph-matching \citep{gong2022multicow}\\
    C) Pose classification using temporal\\
    convolutional network\\
    } &
    Heatmap
    &
    \makecell{Activity\\Classification\\(lying, standing,\\walking)} & 
    \makecell{
        Skeleton extraction method has\\
        high accuracy. Comparisons\\
        were made to find the optimal\\
        model, attention mechanism and\\
        activation function. Final model\\
        displays high performance.
    }\\
    \hline
    \citep{barney2023deep}&
    Dairy &
    \makecell{3000 images\\for pose\\estimation,\\25 videos\\for lameness\\detection} & 250 &
    \makecell{15: Nose, Head, Scapula,\\Withers, Centre of the\\back, Hook bone,\\Tail setting, 8 on\\Legs (Knees, Hooves)} &
    \makecell{Multi Subject\\(maximum 3)\\but IoU=0} &
    \makecell{\textbf{Device}: GoPro Hero 6} &
    Top-Down &
    \makecell{
    A) Tracking using \textbf{SORT} \citep{bewley2016simple}\\
    B) Pose estimation using a \textbf{Mask R-CNN}\\
    C) Feature extraction from pose estimations\\
    (Back arching from 5 keypoints,\\
    head position from 2 keypoints)
    } &
    Regression &
    \makecell{
        Lameness\\Detection \&\\
        Classification
    } &
    \makecell{
        A method with significant\\
        performance in three-fold\\
        lameness detection and\\
        classification.
    } \\
    \hline
    \citep{russello2022t} &
    Dairy &
    \makecell{1059 frame\\groups\\(4 frame\\in each)} &
    30 &
    \makecell{17: Nose, Forehead,\\Withers, Sacrum, Caudal\\thoracic vertebrae, 12 on\\Legs (Carpals or Tersal,\\Fetlocks, Hooves)} & Single Subject &
    \makecell{\textbf{VRS}: 1080p, 30fps, 7.5sec \\ \textbf{View}: Side \\ \textbf{Env}: Left to right walking path} &
    \makecell{Single Subject\\Cropped} &
        \makecell{
        A) Custom \textbf{LEAP} \citep{pereira2019fast}:\\
        One \textbf{poolling-conv} to encoder and\\
        one \textbf{transposedconv-conv} to decoder\\
        B) Dimension added to data for time\\
        C) All convs changed to 3D\\
        Note: Cow cropped in image, with\\
        100px margin on each side
    } &
    \makecell{Heatmaps\\(labels from\\last frame)} &
    \makecell{Pose\\Estimation} &
    \makecell{
        Shows the proposed temporal\\
        model's superiority to static\\
        ones on occluded data and proves\\
        that a deeper architecture is\\
        more beneficial in complicated tasks.
    }  \\
    \hline
    \citep{quek2019automatic} &
    Dairy &
    \makecell{495 images\\for pose\\estimation,\\313 videos\\for lameness\\scoring} &
    70 &
    \makecell{17: DLC model} &
    Single Subject &
    \makecell{\textbf{VRS}: 1080p, 30fps \\ \textbf{CP}: 2m high, 4m away \\ \textbf{View}: Side } &
    Single Subject &
    \makecell{
        A) Keypoint detection and tracking using \textbf{DLC}\\
        B) Feature extraction from poses
    } &
    Heatmaps &
    \makecell{Lameness\\Scoring} &
    \makecell{
        Model's detection certainty in early\\
        stages of lameness is not high, but it\\
        can detect severe lameness with very\\
        high certainty.
    }\\
    \hline
    \citep{li2022basic} &
    Dairy &
    \makecell{
        300 videos\\
        (100 5-15s\\
        videos for\\
        each of\\
        standing,\\
        walking, and\\
        lying\\
        activities)
    } &
    300 &
    \makecell{15: Cow Mouth, Ear,\\Back, 12 on Legs\\(Roots, Knees, Hooves)} &
    Single Subject &
    \makecell{\textbf{VRS}: 6-15sec, \textbf{FOV}: 2.5*cow size \\ \textbf{CP}: 7m away \\ \textbf{View}: Side \\ \textbf{Env}: Left to right walking path} &
    \makecell{Single Subject} &
    \makecell{
        A simple \textbf{HRNet} is used.
    } &
    Heatmap &
    \makecell{Activity\\Classification} & 
    \makecell{
        Proposed a model that can\\
        classify defined actions of\\
        cows with high accuracy.\\
        The model showed slight changes\\
        in accuracy when tested on data\\
        with changed brightness or\\
        added noise.
    } \\
    \hline
    \citep{weng2023video} &
    Dairy &
    300 Videos &
    About 500 &
    \makecell{
        16: head, neck, shoulder\\
        hip, and elbow, knee, and\\
        hoof of each leg
    } &
    \makecell{
        Single and\\
        multi subject
    } &
    \makecell{
        \textbf{Device}: 4*5MP IR camera\\
        3T56WD-I3, 1*4MP 180$^\circ$ wide\\
        camera DS-2CD3T45P1-IS\\
        \textbf{CP}: 3*IR placed higher than cows,\\
        1*IR and 1*wide placed parallel\\
        \textbf{View}: Side
    } &
    Bottom-up &
    \makecell{
        Keypoint detection using a\\
        \textbf{UNet}-like structure\\
        Skeleton estimation using \textbf{PAF}\\
        Tracking algorithm based\\
        on \textbf{Kalman Filtering}\\
        \citep{simon2001kalman}
    } &
    Heatmap &
    \makecell{Pose\\Estimation} &
    \makecell{
        Proposed a model that can\\
        perform pose estimation in\\
        several scenes with single\\
        and multiple object with\\
        high precision
    }\\
    \hline
    \citep{hua2023effective} &
    Dairy &
    \makecell{
        1800 images\\
        for pose\\
        estimation,\\
        400 10s\\
        videos for\\
        activity\\
        classification
    } &
    Not mentioned &
    \makecell{
        17 : Mouth, Forhead, Neck,\\
        Back, Tail, 12 on\\
        Legs (Forelimbs,\\
        Knees, Hooves)
    } &
    Single Subject &
    \makecell{
        \textbf{VRS}: 704*480, 25fps, $10\pm4$sec\\
        \textbf{Env}: Cowshed for lying,\\
        long corridor for standing,\\
        walking, and lameness.\\
        \textbf{Actions}: 1. standing, 2. lying,\\
        3. walking, 4. lameness
    } &
    Single Subject &
    \makecell{
        A) Pose estimation using \textbf{YOLO-pose}\\
        with \textbf{YOLOx} structure\\
        B) Action recognition/classification\\
        using a \textbf{3D-CNN}
    } &
    Heatmaps & 
    \makecell{Pose\\Estimation\\Action\\Classification} &
    \makecell{
        Results show the benefits of\\
        proposed models and their\\
        superiority in comparison to\\
        other models in action classification
    }\\
    \hline
    \citep{khin2022cattle} &
    Dairy &
    4200 images &
    Not mentioned &
    \makecell{
        8: Head, Mid-body,\\
        base of tail, end\\
        of tail, one for\\
        each leg
    } &
    \makecell{
        Multi Subject\\
        but IoU is\\
        not mentioned
    } &
    \makecell{
        \textbf{VRS}: 1fps, 5min\\
        \textbf{View}: Top\\
        \textbf{Actions}: 1. standing, 2. sitting\\
        3. sitting w/ leg extend 4. eating\\
        5. drinking 6. tail raised
    } &
    Top Down &
    \makecell{
        A) Body part location extraction\\
        using \textbf{DLC} defaults\\
        B) Action classification using\\
        \textbf{SVM} based on extracted locations,\\
        width, and height
    } &
    \makecell{Model is\\not Trained} &
    \makecell{Activity\\Classification} &
    \makecell{
        Width and height of cow bounding\\
        box, and features derived from\\
        extracted keypoints are used\\
        to classify actions. The proposed\\
        algorithm has a high accuracy.       
    }\\
    \hline
    \citep{yang2023scirnet} &
    \makecell{ \begin{turn}{90} \;Not mentioned\; \end{turn} } &
    \makecell{
        AP-10k\\
        for pose\\
        estimation,\\
        100 videos\\
        (2-3s) for\\
        activity\\
        classification
    } &
    40 &
    \makecell{17: AP-10K} &
    \makecell{There are interactions\\and overlaps, but\\IoU is not mentioned} &
    \makecell{
        \textbf{Env}: Pasture\\
        \textbf{Actions}: 1. riding 2. grazing\\
        3. lying 4. walking\\
        \textbf{Interactions}: 1. conflicting\\
        2. mounting 3. smelling 4. licking
    } &
    Top Down &
    \makecell{
        A) Subject detection using \textbf{YOLOx}\\
        B) Pose estimation using \textbf{HRNet}\\
        D) Tracking using \textbf{DeepSort}\\
        \citep{wojke2017simple}\\
        E) Activity recognition using graphs\\
        of inter-body and intera-body poses
    } &
    Heatmap &
    \makecell{Activity\\Classification} &
    \makecell{
        The model focuses on classifying\\
        interactions in addition to\\
        individual actions
    } \\
    \hline
    \citep{taghavi2023cow} &
    Dairy &
    \makecell{758 groups\\of frames\\(2 frame)} &
    34 &
    \makecell{17: Nose, Forehead,\\Withers, Sacrum, Caudal\\thoracic vertebrae, 12 on\\Legs (Carpals or Tersal,\\Fetlocks, Hooves)} &
    \makecell{Multi Subject\\but IoU is\\not mentioned}&
    \makecell{\textbf{Device}: RGB camera\\HikVision IP camera model,\\DS-2CD2T45FWD-I5\\\textbf{VRS}: 1520*2688, 25fps,\\1hr/3hr daily total \\\textbf{CP}: 2.2m high, 4m away\\\textbf{Env}: Milking parlor exit lane,\\Walking back in single line} &
    \makecell{Single Subject\\Cropped} &
    \makecell{
        + Cow detection using \textbf{YOLOv3}\\
        + Data increase by randomly changing\\
        rotation, brightness, and contrast\\
        + Pose estimation using \textbf{T-LEAP}\\
        \citep{russello2022t}
    } &
    \makecell{Heatmap\\(labels from\\last frame)} &
    \makecell{Pose\\Estimation} &
    \makecell{
        Reports the performance of proposed\\
        model and marks steps required to\\
        calculate gait features from\\
        detected keypoints
    } \\
    \hline
    \citep{li2022fusion} &
    Dairy &
    \makecell{680 Videos\\(6s to 15s)} &
    680 &
    \makecell{17: Mouth, Left Eye,\\Right Eye, Left Ear,\\Right Ear, 12 on\\Legs (Roots, Knees,\\Hooves)} &
    Single Subject &
    \makecell{
        \textbf{Device}: DS-2DM1-714\\
        Dome Cameras\\
        \textbf{VRS}: 704*576, 25fps (PAL)\\
        704*480, 30fps (NTSC), 6-15sec\\
        \textbf{FOV}: 2.5*cow length\\
        \textbf{CP}: 7m away and parallel
    } &
    Bottom-Up &
    \makecell{
        + Skeleton estimation using \textbf{VGG-19}\\
        and matching predicted Keypoints and PAFs\\
        + Scoring lameness based on normalized\\
        output of three branches:\\
        1. Multiple \textbf{CNN}s from RGB images\\
        2. A \textbf{ST-GCN} from skeleton data\\
        3. Multiple \textbf{CNN}s from Flow
    } &
    Heatmaps &
    \makecell{Lameness\\Scoring} &
    \makecell{
        The optimal weighs in combining\\
        the branches of the proposed model\\
        are described and the model is \\
        reported when testing on data\\
        with changes in illumination
    } \\
    \hline
    \citep{islam2023analysis} &
    Beef &
    78 Videos &
    4 &
    \makecell{5: Head, Upper neck,\\Lower neck, Left\\ear, Right ear} &
    Single Subject &
    \makecell{\textbf{Device}: RPi camera (8MP IMX219)\\\textbf{View}: Two viewpoints\\\textbf{Notes}: Two ultrasonic sensors,\\for triggering camera\\and as ground truth}  &
    Single Subject &
    \makecell{
        + Keypoint Detection using \textbf{DLC}\\
        + Detecting drinking behaviour\\
        using \textbf{LSTM}
    } &
    Heatmaps &
    \makecell{Drinking\\Detection} &
    \makecell{
        Drinking behaviour and drinking\\
        time are predicted with high\\
        accuracy in the proposed method.\\
        The employed camera system can be\\
        used in feedlots. 
    } \\
    \hline
    \citep{gardenier2018object} &
    Dairy &
    \makecell{
        74\\
        near infrared\\
        images
    } &
    223 &
    \makecell{
       8: Carpal/tarsal joints\\
       and hooves 
    } &
    Single Subject &
    \makecell{
        \textbf{Device}: 4*Kinect-v2 Sensors\\
        \textbf{VRS}: 512*424, 30fps\\
        \textbf{CP}: Left side: 0.6m high,\\
        1.5m away, top: 2.7m high\\
        \textbf{View}: Left side and top\\
        \textbf{Env}: Walking at exit of\\
        rotary milker
    } & 
    Single Subject &
    \makecell{
        Boyd part detection using\\
        \textbf{Faster R-CNN} \citep{ren2015faster}\\
        with \textbf{VGG16} 
    } &
    Regressor &
    \makecell{Pose\\Estimation} &
    \makecell{
        Proposed a method for efficient\\
        keypoint detection and tracking.\\
        Analysis is done on tracking\\
        results to specify measures for\\
        detecting lameness such as rear\\
        hoof placement pattern.
    } \\
    \hline
    \citep{jiang2019flyolov3} &
    Dairy &
    \makecell{
        800 images\\
        from recordings\\
        200 images\\
        by adding noise
    } &
    Not mentioned &
    \makecell{
        6: Head, trunk and\\
        four legs
    } &
    Single Subject &
    \makecell{
        \textbf{Device}: SONY HDR-CX290E\\
        and a webcam\\
        \textbf{VRS}: 25fps\\
        \textbf{CP}: Perpendicular to cow's\\
        walking direction\\
    } &
    Single Subject &
    \makecell{
        Body part detection with:\\
        + \textbf{FilterLayer YOLOv3} and\\
        \textbf{YOLOv3} on \textbf{Darknet}\\
        + \textbf{Faster R-CNN} on \textbf{Caffe}
    } &
    Regressor &
    \makecell{Body Part\\Segmentation} &
    \makecell{
        Accuracy, recall rate, average\\
        frame rate, and average\\
        accuracy are reported for\\
        comparing the three networks.
    } \\
    \hline
    \citep{okour2022sim2real}&
    \makecell{ \begin{turn}{90} \;Not applicable\; \end{turn} }&
    \makecell{
        1200 samples\\
        from a 40sec\\
        animation
    } &
    1 (Synthetic)&
    \makecell{
       13: Three at each front leg,\\
       two at each back leg,\\
       two at hip bone,\\
       one at front of spine
    } &
    Single Subject &
    \makecell{
        \textbf{Device}: 17 simulated\\
        depth cameras\\
        \textbf{CP}: Eight on each\\
        side, one at top
    } &
    Single Subject &
    \makecell{
        Keypoint detection with\\
        \textbf{Pointnet++} \citep{qi2017pointnet++}
    } &
    Heatmap &
    \makecell{Pose\\Estimation} &
    \makecell{
        Model compared to a model\\
        trained on real data,\\
        real data used for test
    } \\
    \hline
    \citep{wang2023op} &
    \makecell{ \begin{turn}{90} \;Not mentioned\; \end{turn} }&
    \makecell{
        Not\\
        mentioned
    } &
    Not mentioned &
    \makecell{
        20: Eyes, ears, snout,\\
        two on each leg,two on\\
        back, and five mid-body
    } &
    Multi Subject &
    Not mentioned &
    Bottom-up &
    \makecell{
        Skeleton estimation using\\
        \textbf{Mask R-CCn} with \textbf{PAF}s
    } &
    Heatmap &
    \makecell{
        Cattle\\Identification
    } &
    \makecell{
        Proposed model has significantly\\
        reduced training time and improved\\
        accuracy and can be employed in\\
        complex complex environments. 
    } \\
    \hline
    \citep{liu2023feature} &
    Dairy &
    505 videos &
    196 &
    Same as \citep{liu2020cow} &
    Single Subject &
    \makecell{
        \textbf{Device}: 2*Ubiquiti Networks\\
        UniFi UVC G3-FLEX\\
        \textbf{VRS}: 1080p, 30fps\\
        \textbf{CP}: 3m high, front:\\
        4.3m away, side: 7.9m away\\
        \textbf{View}: Front and side\\
        \textbf{Env}: Walking single file
    } &
    \multicolumn{3}{c|}{
        Same as \citep{liu2020cow} 
    } &
    \makecell{
        Weight\\Estimation
    } &
    \makecell{
        Weight\\Estimation
    } \\
    \hline
    \citep{li2024automated} &
    Beef &
    2956 images &
    30 &
    \makecell{
        8: Back, front legs,\\
        throat, and hip
    } &
    \makecell{
        Single Subject\\
        (unanotated subjects\\
        may be present\\
        in background)
    } &
    \makecell{
        \textbf{Device}: Wide-angle camera\\
        \textbf{CP}: 0.8m high, 1m-5m away\\
        \textbf{View}: Side
    } &
    Top-Down &
    \makecell{
        + Cow detection with \textbf{YOLOv5}\\
        + Keypoint detection with\\
        \textbf{Lite-HRNet}\\
        \citep{yu2021lite}
    } &
    Heatmap &
    \makecell{
        Body Size\\Measuring
    } &
    \makecell{
        Detected keypoints combined\\
        with a depth estimation network\\
        to allow measurement of body pars.\\
        Steps required to reach the desired\\
        robustness are pointed out. 
    } \\
    \hline
\end{longtable}
\end{landscape}
\pagebreak

\end{document}